\documentclass{article}
\usepackage{proceed2e}

\usepackage{bm}
\RequirePackage{amsthm,amsmath,natbib,amssymb}
\usepackage{algorithm, algorithmic}
\usepackage{epsfig}
\usepackage{subfigure}

\theoremstyle{plain}
\newtheorem{thm}{Theorem}
\newtheorem{lem}[thm]{Lemma}
\newtheorem{definition}[thm]{Definition}

\newcommand{\fracless}[2]{\genfrac{}{}{0pt}{}{#1}{#2}}

\DeclareMathOperator*{\argmax}{argmax}

\DeclareMathOperator*{\LOCAL}{LOCAL}
\DeclareMathOperator*{\OUT}{OUT}
\DeclareMathOperator*{\CUT}{CUT}
\DeclareMathOperator*{\SDEF}{SDEF}
\DeclareMathOperator*{\tr}{Tr}
\DeclareMathOperator*{\diag}{diag}
\newcommand{\norm}[1]{\left|\left|#1\right|\right|}

  \renewenvironment{thebibliography}[1]{%
    \begin{oldthebibliography}{#1}%
      \setlength{\parskip}{0ex}%
      \setlength{\itemsep}{0ex}%
  }%
  {%
    \end{oldthebibliography}%
  }

\title{Improved Estimation of High-dimensional Ising Models} 
 
\author{ {\bf Mladen Kolar}\\
School of Computer Science\\
Carnegie Mellon University\\
Pittsburgh, PA 15213\\
mladenk@cs.cmu.edu
\And
{\bf Eric P. Xing}\\
School of Computer Science\\
Carnegie Mellon University\\
Pittsburgh, PA 15213\\
epxing@cs.cmu.edu
} 
 
\begin{document} 
 
\maketitle

\begin{abstract} 
We consider the problem of jointly estimating the parameters as well
as the structure of binary valued Markov Random Fields, in contrast to
earlier work that focus on one of the two problems. We formulate the
problem as a maximization of $\ell_1$-regularized surrogate likelihood
that allows us to find a sparse solution. Our optimization technique
efficiently incorporates the cutting-plane algorithm in order to
obtain a tighter outer bound on the marginal polytope, which results
in improvement of both parameter estimates and approximation to
marginals.  On synthetic data, we compare our algorithm on the two
estimation tasks to the other existing methods. We analyze the method
in the high-dimensional setting, where the number of dimensions $p$ is
allowed to grow with the number of observations $n$.  The rate of
convergence of the estimate is demonstrated to depend explicitly on
the sparsity of the underlying graph.
\end{abstract} 

\section{Introduction}

Undirected graphical models, also known as Markov random fields
(MRFs), have been successfully applied in a variety of domains,
including natural language processing, computer vision, image
analysis, spatial data analysis and statistical physics. In most of
these domains, the structure of the graphical models is constructed by
hand. However, in certain complex domains we have little expertise
about interactions between features in data and we need a method that
automatically selects a model that represents data well. We propose an
algorithm that is able to learn a sparse model that fits data well and
has an easily interpretable structure.

Let $X = (X_1,\ldots, X_p)^T$ be a random vector with distribution $P$
that can be represented by an undirected graph $G = (V, E)$. Each
vertex from the set $V$ is associated with one component of the random
vector $X$. The edge set $E$ of the graph $G$ encodes certain
conditional independence assumptions among subsets of the
$p$-dimensional random vector $X$; $X_i$ is conditionally independent
of $X_j$ given the other variables if $(i, j) \notin E$. Let
$\mathcal{D} = \{x^{(i)} = (x_1^{(i)}, \ldots, x_p^{(i)})\}_{i=1}^n$
be an i.i.d. sample. Our task is to estimate the edge set $E$ as well
as the parameters of the distribution that generated the sample. The
main contribution of our paper is twofold: the development of an
efficient algorithm that estimates the undirected graphical model from
data, and the high-dimensional asymptotic analysis of its
estimates. We find that the rate of convergence of the estimate
explicitly depends on the sparsity of the underlying graphical model.

Under the assumption that $X$ is Gaussian, estimation of the graph
structure is equivalent to estimation of zeros in the inverse
covariance matrix $\Sigma^{-1}$. Several methods have been proposed
that estimate the graph structure from
$\mathcal{D}$. \cite{Drton04model} proposed a method that tests for
partial correlations that are not significantly different from zero,
which can be applied when $p$ is small. In high dimensional setting,
when $p$ is large, the estimation is more complicated, but under the
assumption that the graph is sparse, several methods can be employed
successfully for structure recovery (e.g \cite{meinshausen06high,
  banerjee07model, Friedman07glasso, rothman08sparse}). If the random
variable $X$ is discrete, the problem of structure estimation becomes
even harder as the likelihood cannot be optimized efficiently due to
the problem of evaluation of the log-partition function. Many recently
proposed methods make use of $\ell_1$ regularization to learn sparse
undirected models. \cite{Wainwright06High} used a pseudo-likelihood
approach, based on the local conditional likelihood at each node,
which results in a consistent estimate of the structure, but not
necessarily of the parameters. \cite{lee07efficient} proposed to
optimize the $\ell_1$ penalized log-likelihood only over the set of
active variables and iteratively enlarge the set until optimality is
achieved. However, the resulting graph structure is not necessarily
sparse. \cite{banerjee07model} used the log-determinant relaxation of
the log-partition function \cite{wainwright06log} to obtain a
surrogate likelihood that can be easily optimized.

In this paper we are interested in learning both the parameters and
the structure of a binary valued Markov Random Field with pairwise
interactions from observed data. An important insight into the problem
of structure estimation is that even if a sparse graph is estimated,
that does not necessarily mean that the inference in the model is
tractable (e.g. inference in a grid, in which each node has only 4
neighbors, is not tractable). Since we are interested in using the
estimated model for inference, we use the insight obtained from
\cite{wainwright06wrong}; i.e.~we use the same approximate procedure
for estimating the parameters and for inference, as the bias
introduced in estimation phase can compensate for the error in the
inference phase. Our method is mostly related to
\cite{banerjee07model}, as we propose to optimize the $\ell_1$
penalized surrogate likelihood based on the log-determinant
relaxation. However, in order to obtain a better estimate, we
efficiently incorporate the cutting-plane algorithm
\cite{sontag08bounds} for obtaining a tighter outer bound on the
marginal polytope into our optimization procedure. Using a better
approximation to the log-partition function is crucial in reducing the
approximation error of the estimate. In many applications, the ambient
dimensionality of the model $p$ is larger than the sample size $n$ and
the classical asymptotic analysis, where the model is fixed and the
sample size increases, does not give a good insight into the model
behavior. We analyze our estimate in the high-dimensional setting,
i.e. we allow the dimension $p$ to increase with the sample size
$n$. Doing so, we are allowing a procedure to select a more complex
model that can represent a larger class of distributions.

\section{Preliminaries}

In this section we briefly introduce MRFs. We present why the exact
inference in discrete MRFs, in general, is intractable, and how to
formulate approximate inference as a convex optimization problem. In
Section \ref{sec:structure}, we derive our learning algorithm
using methods presented here.

\textbf{Markov Random Fields. } Consider an undirected graph $G$ with
vertex set $V = \{1, 2, \ldots, p\}$ and edge set $E \subseteq V \times
V$. A Markov random field consists of a random vector $X =
(X_1,\ldots, X_p)^T \in \mathcal{X}^p$, where the random variable
$X_s$ is associated with vertex $s \in V$. In this paper we will
consider binary pairwise MRFs, the Ising model, in which the
probability distribution factorizes as $\mathbb{P}(x) = \exp \left(
\langle \theta, \varphi(x) \rangle - A(\theta) \right)$ and $X \in
\{-1, 1\}^p$. Here $\langle \theta, \varphi(x) \rangle$ denotes the
dot product between the parameter vector $\theta \in \mathbb{R}^d$ and
potentials $\varphi(x) \in \mathbb{R}^d$, and $A(\theta) = \log
\sum_{x\in\mathcal{X}^p} \exp \left( \langle\theta,
\varphi(\bm{x})\rangle \right)$ is the log-partition function. Since
we are considering binary pairwise MRFs, potentials are functions over
nodes and edges of the form $\varphi(x_v) = x_v$ and $\varphi(x_u,
x_v) = x_ux_v$. For future use, we introduce a shorthand notation
for the mean parameters $\eta_v = E_\theta[\phi(x_v)]$ and $\eta_{vv'}
= E_\theta[\phi(x_v,x_{v'})]$.

\textbf{Log-partition function. } Evaluating the log-partition
function involves summing over exponentially many terms and, in
general, is intractable. The log-partition function can be expressed
as an optimization problem, as a variational formulation, using its
Fenchel-Legendre conjugate dual $A^*(\eta)$:
\begin{equation} \label{eq:logpart_dual}
A(\theta) =
\sup_{\eta\in\mathcal{M}}\{\langle\theta,\eta\rangle-A^*(\eta)\},
\end{equation} 
where $\mathcal{M} := \left\{ \eta\in\mathbb{R}^d \ |\ \exists
p(\bm{X}) \text{ s.t. } \eta = E_\theta[\phi(\bm{x})] \right\}$ is the
set of realizable mean parameters $\eta$ and the dual function
$A^*(\eta) = -H(p(x;\eta(\theta))$ is equal to the negative entropy of
the distribution parametrized by the mean parameters $\eta$. For each
parameter $\theta$ there is a corresponding mean parameter
$\eta\in\mathcal{M}$ that maximizes \eqref{eq:logpart_dual}. The
relation is given as:
\begin{equation} \label{eq:mean_params}
\eta = \nabla A(\theta) = E_\theta[\phi(\bm{x})].
\end{equation}
\cite{wainwright03graphical, wainwright06log} list other properties of
log-partition function.

\textbf{Log-partition relaxations. } The log-partition function
written in equation \eqref{eq:logpart_dual} defines an optimization
problem restricted to the set $\mathcal{M}$. Since the set
$\mathcal{M}$ is a polytope, it can be represented as an intersection
of a finite number of hyperplanes; however, the number of hyperplanes
needed to describe the set $\mathcal{M}$ grows exponentially with the
number of nodes $p$. It is important to find an outer bound on
$\mathcal{M}$ that can be easily characterized and as tight as
possible. One outer bound can be obtained using the set of points that
satisfy local consistency conditions
$
\LOCAL(G):=$
$$ \left\{ \eta \in [-1,1]^d \ \Big|
\ \forall(u,v) \in E\ :\ 
\begin{array}{l}
\eta_{uv} - \eta_{u} + \eta_{v} \leq 1\\ 
\eta_{uv} - \eta_{v} + \eta_{u} \leq 1\\
\eta_{u} + \eta_{v} - \eta_{uv} \leq 1 
\end{array}
\right\}.
$$
By construction, we have $\mathcal{M} \subseteq \LOCAL(G)$, but points
in $\LOCAL(G)$ do not necessarily represent mean parameters of any
probability distribution. Another outer bound can be obtained
observing that the second moment matrix $M_1(\eta) = E_\theta
[(1\ \bm{x})^T (1\ \bm{x})]$ is positive semi-definite. Therefore we
have $\mathcal{M} \subseteq \SDEF_1(G) := \left\{ \eta \in
\mathbb{R}^d\ |\ M_1(\eta) \succeq 0 \right\}$.

The outer bound on the polytope $\mathcal{M}$ can be further tightened
by relating it to the cut polytope $\CUT(G)$ (e.g. \cite{deza97}) and
using known relaxations to the cut polytope. Mapping between
$\mathcal{M}$ and the cut polytope can be done using the
\textit{suspension graph} $G'=(V', E')$ of the graph $G$, where $V' =
V \cup \{p+1\}$ and $E' = E \cup \{(v,p+1)\ |\ v \in V \}$. The
suspension graph is created by adding an additional node $p+1$ to the
graph G and connecting each node $v \in V$ to the newly created node
$p+1$.
\begin{definition}
Let $w_{uv}$ denote the weight of an edge in $G'$. The linear
bijection $\xi_{cut}$ that maps points $\eta \in \mathcal{M}$ to
points $w \in \CUT(G')$ is given by $w_{v,n+1} =
\frac{1}{2}(\eta_v+1)$ for $v \in V$ and $w_{uv} = \frac{1}{2}
(1-\eta_{uv})$ for $(u,v) \in E$.
\end{definition}
Using a separation algorithm, it is possible to separate a class of
inequalities that define the cut polytope and add them to the
inequalities defining $\LOCAL(G)$ to tighten the outer
bound. Efficient separation algorithms are known for several classes
of inequalities \cite{deza97}, but in this paper we will use the
simplest one, \textit{cycle-inequalities}. Cycle-inequalities can be
written as:
\begin{equation} \label{eq:cycle}
\sum_{uv \in C\backslash F}w_{uv} + \sum_{uv \in F}(1-w_{uv}) \geq 1
\end{equation}
where $C$ is a cycle in $G'$ and $F \subseteq C$ and $|F|$ is odd. The
class of cycle-inequalities can be efficiently separated using
Dijkstra's shortest path algorithm in $\mathcal{O}(n^2\log n +
n|E|)$. 

To further relax the problem \eqref{eq:logpart_dual}, we approximate
the negative entropy $A^*(\eta)$ by a function $B^*(\eta)$ to obtain
the approximation $B(\theta)$ to $A(\theta)$:
\begin{equation} \label{eq:approx_logdet}
B(\theta) =
\sup_{\eta\in\OUT(G)}\ \langle\theta,\eta\rangle-B^*(\eta),
\end{equation}
where $\OUT(G)$ is a convex and compact set acting as an outer bound
to $\mathcal{M}$. Even though it is not required, many of the existing
entropy approximations are strictly convex (e.g. the convexified Bethe
entropy \cite{wainwright05upper}, the Gaussian-based log-determinant
relaxation \cite{wainwright06log} or the reweighted Kikuchi
approximations \cite{wiegerinck05}), which guarantees uniqueness of
the solution to \eqref{eq:approx_logdet}.

\textbf{Inference in MRFs. } The inference task in MRFs refers to
finding or approximating the marginal probabilities (or the mean
vector). It can be seen from equation \eqref{eq:mean_params} that the
log-partition plays an important role in inference and a good
approximation to it is essential in obtaining good estimates of the
marginals. Many known approximate inference algorithms can be
explained using the framework explained above; they create an outer
bound $\OUT(G)$ and use an entropy approximation to estimate
the marginals (e.g. the log-determinant relaxation
\cite{wainwright06log}, Belief Propagation \cite{Yedidia05} and
tree-reweighted sum-product (TRW) \cite{wainwright05upper} to name
few). Recent work proposed a cutting-plane algorithm
\cite{sontag08bounds} that iteratively tightens the outer bound on the
polytope $\mathcal{M}$ using cycle-inequalities and empirically
obtains improved estimates of marginals.

\section{Structure learning and parameter estimation} \label{sec:structure}

In this section we address the problem of structure learning and
parameter estimation. Given a sample $\mathcal{D}$, we obtain our
estimate of the edge set $E$ and parameters $\theta$ associated with
edges as a maximizer of the $\ell_1$ penalized surrogate
log-likelihood:
\begin{equation} \label{eq:mle_conv}
\begin{aligned}
\hat{\theta}^n &= \argmax_{ \theta \in \mathbb{R}^d} \ \ell(\theta;
\mathcal{D}) - \lambda^n \sum_{uv \in E}|\theta_{uv}| \\ 
&= \argmax_{\theta \in \mathbb{R}^d} \ \langle \theta, \hat{\eta}^n \rangle -
B(\theta) - \lambda^n \sum_{uv\in E}|\theta_{uv}|,
\end{aligned}
\end{equation}
where $\hat{\eta}^n$ denotes the mean parameter estimated from the
sample. $\lambda^n$ is a tuning parameter that sets the strength of
the penalty. Note that a solution to the problem \eqref{eq:mle_conv}
simultaneously gives the estimate of the edge set $E$ and the
parameter vector $\theta$, since if $\hat\theta_{uv} = 0$ then the
estimated graph does not contain the edge between nodes $u$ and
$v$. Since the $\ell_1$ penalty shrinks the edge parameters towards
$0$, the estimation is biased towards sparse models.

In the remainder of this section we will explain our algorithm that
efficiently solves \eqref{eq:mle_conv}.  Since the problem
\eqref{eq:mle_conv} is convex one could use, for example, the
subgradient method for non-diferentiable functions
\cite{bertsekas99}. However, computing the gradient of the
log-likelihood $\frac{\partial \ell(\mathcal{D}; \theta)}{\partial
  \theta} = \hat\eta^n - E_\theta[\varphi(x)]$ requires inference over
the model with current values of the parameters. Since the inference
is only approximate, the accuracy of the computed gradient heavily
depends on the approximation used. Note again that the sparsity of the
graph does not necessarily imply that the inference is tractable.
Applying the cutting-plane algorithm \cite{sontag08bounds} for
approximate inference could achieve good accuracy, however it would be
computationally prohibitive, since at each iteration of the
subgradient algorithm, the cutting plane algorithm have to be run anew
to obtain the mean parameters. Hence, computational inefficiency stems
from the fact that for each parameter $\theta$, we have to compute the
corresponding mean parameter $\eta$.  We present a way to exploit the
structure of the log-determinant relaxation to obtain both $\theta$
and $\eta$ and obtain computationally efficient algorithm.  We will
use the convenient way of representing parameters in a matrix:
\begin{equation*}
R(\theta) = \left(
\begin{matrix}
0 & \theta_1 & \theta_2 & \ldots & \theta_p \\
\theta_1 & 0 & \theta_{12} & \ldots & \theta_{1p} \\
\vdots & & & & \\
\theta_p & \theta_{1p} & \theta_{2p} & \ldots & 0 \\
\end{matrix}
\right). 
\end{equation*}

\begin{algorithm}[t]
\caption{Structure learning with cutting-plane
  algorithm} \label{alg:struct_learn}
\begin{algorithmic}[1]
\STATE {$\OUT(G) \leftarrow \LOCAL(G)$}
\REPEAT
\STATE { $ \hat\theta^n, \hat\eta \leftarrow 
  \max_{\theta} \{ {\small
\begin{array}{c}
  \langle \theta, \hat\eta^n \rangle -
  \lambda^n\sum_{vv'}|\theta_{vv'}| - \\
  \max_{\eta \in \OUT(G)}\ \{
  \langle \theta, \eta \rangle - B^*(\eta) \} 
\end{array} } \}$ }
\STATE { $\bm{w} \leftarrow \mathrm{Create\_Suspension\_Graph}(\hat\eta)$ }
\STATE { $\mathcal{C} \leftarrow
  \mathrm{Separate\_Cycle\_Inequalities}(\bm{w}) $}
\STATE { $ \OUT(G) \leftarrow \OUT(G) \cap \mathcal{C} $}
\UNTIL{ $\mathcal{C} = \emptyset$ }
\end{algorithmic}
\end{algorithm}

Algorithm \ref{alg:struct_learn} describes our proposed method for
structure learning. To obtain the solution of \eqref{eq:mle_conv}
efficiently, we use variational representation of $B(\theta)$ using
the log-determinant relaxation and jointly optimize over $\theta$ and
$\eta$. Starting with $\LOCAL(G)$ as an outer bound to $\mathcal{M}$,
the algorithm alternates between finding the best parameters and
tightening the outer bound $OUT(G)$ by incorporating the cycle
inequalities that are violated by the current mean parameters
$\eta$. The algorithm is similar, in spirit, to the cutting-plane
algorithm \cite{sontag08bounds} for inference. However, optimization
in line 3 of Algorithm \ref{alg:struct_learn} is done over all
parameters jointly, which produces some technical challenges. We
proceed with a procedure for solving the optimization problem
\eqref{eq:mle_conv}.

 The formulation in line 3 arose from using the variational form
 \eqref{eq:approx_logdet} of $B(\theta)$ in the surrogate likelihood
 \eqref{eq:mle_conv}. The idea behind the log-determinant relaxation
 \cite{wainwright06log} is to upper bound the log-partition function
 using the Gaussian-based entropy approximation: 
$$ A(\theta) \leq \sup_{\eta \in \OUT(G)}
 \frac{1}{2}\log\det(R(\eta)+\diag(m)) + \langle \theta, \eta \rangle,
$$
where $m\in\mathbb{R}^{p+1} = (1,\frac{4}{3},\ldots,\frac{4}{3})$. To
make use of this upper bound, we have to rewrite it so that both
parameters $\theta$ and $\eta$ can be extracted from it. Before we
rewrite the upper bound, notice that after $k$ iterations of
repeat-until loop in Algorithm \ref{alg:struct_learn}, we have added
$k$ cycle-inequalities. It will be useful to rewrite equation
\eqref{eq:cycle} in a matrix form as $\bm{\tr}(A_kR(\eta)) \geq b_k$,
where $A_k$ is a symmetric matrix for the $k$-th inequality.

\begin{lem} \label{lem:relax_logpart}
After $k$ iterations of algorithm, we write the log-partition
relaxation as
\begin{equation}
{\small
\begin{aligned}
B(\theta) &= \frac{p}{2}\log(\frac{\mathrm{e}\pi}{2})
-\frac{1}{2}(p+1)
- \frac{1}{2} \max_{\nu, \alpha \geq 0} \{ \nu^Tm 
- \alpha^Tb \\
& + \log\det(-R(\theta)-\diag(\nu) + \sum_{i=1}^{k}\alpha_iA_i) \}.
\end{aligned}
}
\end{equation}
To obtain the mean vector $\eta^*$ corresponding to $\theta$ we take
off diagonal elements of the matrix
$$ Z = (-R(\theta) - \diag(\nu) + \sum_i\alpha_iA_i)^{-1}
$$ 
defined for optimal $\nu, \alpha$, i.e. $R(\eta^*) = Z - \diag(Z)$.
\end{lem}

\begin{proof}
Due to the lack of space, we leave out technical details of the proof
and just give a sketch of the main idea. The proof is similar to the
proof of Lemma 5 in \cite{banerjee07model}. We start from equation
\eqref{eq:approx_logdet}, where the Gaussian-based entropy is used as
$B^*(\eta)$, and rewrite it in the Lagrangian form with $\alpha_i$ as
a Lagrangian multiplier for $i$-th cycle-inequality. The lemma follows
from rewriting the Lagrangian in the dual form.
\end{proof}

Using Lemma \ref{lem:relax_logpart}, we can rewrite the problem in
line 3 of Algorithm \ref{alg:struct_learn} so that both $\theta$ and
$\eta$ can be extracted. Defining $Y := -R(\theta) - \diag(\nu)$ and
dropping constant terms the optimization problem is written as:
\begin{equation} \label{eq:ml_cycle}
{\small
\max_{\fracless{Y}{\alpha \geq 0}} \left\{  
\begin{array}{c}
- \bm{\tr}(Y(R(\hat\eta^n) + \diag(m))) + \\
\log\det \left(Y + \sum_i \alpha_iA_i \right) -
\alpha^Tb - \lambda^n\sum_{uv}|Y_{uv}| 
\end{array}
\right\}. }
\end{equation}
With $\alpha = 0$, the problem \eqref{eq:ml_cycle} is identical to the
problem analysed in \cite{banerjee07model, Friedman07glasso}, where
$Y$ can be found using a block coordinate descent. However, due to the
terms that arise from the added cycle-inequalities, the Lagrangian
multipliers $\alpha \neq 0$ are different from zero and we need a
different method to obtain the optimal $Y$ and $\alpha$.

\begin{algorithm}[t]
\caption{Finding best parameters} \label{alg:solve_ml}
\begin{algorithmic}[1]
\STATE { $W, \alpha \leftarrow \text{Initialize()}$ }
\REPEAT
\STATE { ${\small W \leftarrow \max_W \left\{
  \begin{array}{c} 
    \log\det(W+R(\hat\eta^n)+\diag(m)) \\
    - \bm{\tr}(W(\sum_i\alpha_iA_i)) 
  \end{array} \right\} }$ \\
  $\quad W \in \mathcal{W}$ ( defined in equation \eqref{eq:box_constraint} )
} \label{alg:box_constr}
\STATE { $ Y \leftarrow (W + R(\hat\eta^n) + \diag(m))^{-1} -
  \sum_i\alpha_iA_i $ }
\STATE { $ \alpha \leftarrow \max_{\alpha\geq0}\ \{ \log\det(Y +
  \sum_i\alpha_iA_i) - \alpha^Tb\} $}
\UNTIL{ stop criterion }
\end{algorithmic}
\end{algorithm}

We propose Algorithm \ref{alg:solve_ml} for solving the problem
\eqref{eq:ml_cycle}.  The algorithm iterate between solving $Y$ and
$\alpha$.  For fixed $\alpha$, the dual of \eqref{eq:ml_cycle} is
given as
\begin{equation} 
  \max_{W \in \mathcal{W}} \left\{
  \begin{array}{c}
    \log\det (W+R(\hat\eta^n)+\diag(m)) - \\
    \bm{\tr}(W(\sum_i\alpha_iA_i))
  \end{array}
  \right\},
\end{equation}
where 
\begin{equation} \label{eq:box_constraint}
{ \small
\mathcal{W} = \left\{ 
\begin{array}{ll}
     W_{uv} = 0 & \text{for } u=1,\ v=1,\ u = v\\
     |W_{uv}| \leq \lambda^n & \text{otherwise }
\end{array}  \right\}. }
\end{equation}
To solve for $Y$ we apply the subgradient method, in which we optimize
the dual variable $W$ with the projection step onto the box constraint
defined in \eqref{eq:box_constraint}.  The gradient direction is given
as $\nabla W = (R(\hat\eta^n) + \diag(m) + W)^{-1} -
\sum_i\alpha_iA_i$ and the step size can be defined as $\gamma_t =
\gamma_0/\sqrt{t}$. For fixed $Y$, the problem \eqref{eq:ml_cycle}
reduces to one in line 5 of Algorithm \ref{alg:solve_ml}. Since the
dimension of $\alpha$, corresponding to the number of
cycle-inequalities, is not large, we can apply any optimization
method to find an optimal $\alpha$.

Algorithm \ref{alg:solve_ml} is guaranteed to converge, which can be
shown from the standard results for block coordinate optimization (e.g
\cite{tseng01}). Using the same results, we could perform the analysis
of the convergence rate, but that is beyond the scope of this
paper. The computational complexity of Algorithm \ref{alg:solve_ml} is
roughly $\mathcal{O}(p^3)$, so the overall complexity of the structure
learning is $\mathcal{O}(p^3)$. The method has lower complexity than
the method \cite{banerjee07model} and same as {\it graphical lasso}
\cite{Friedman07glasso}. A useful trick for boosting the performance
of the structure learning algorithm is to use a warm-start strategy
for Algorithm \ref{alg:solve_ml}, i.e.~initialize $Y$ and $\alpha$ to
the optimal solution of the previous iteration.

An important advantage of our algorithm, over other block-coordinate
descent algorithms \cite{banerjee07model, Friedman07glasso}, is that
we can readily modify it to the case when the regularization is given
as a constraint on $\ell_1$ ball in which parameters lie,
e.g. $\sum_{uv\in E}|\theta_{uv}| \leq C$. In that case, we can solve
directly the primal problem, to obtain $Y$, using an efficient
algorithm for projection onto the $\ell_1$ ball \cite{duchi08}. The
algorithm for efficient projection onto the $\ell_1$ ball was also
exploited for learning sparse inverse covariance matrices under the
Gaussian assumption on data \cite{Duchi08projected}. Finally, the
structure learning algorithm could be extended to the non-binary MRFs
using a generalization to the log-determinant relaxation
\cite{wainwright06log} and projecting the marginal polytope to
different binary marginal polytopes \cite{sontag08bounds}.

\section{Asymptotic analysis}

In this section, we state our main theoretical result on the
convergence rate of the $\ell_1$ penalized log-likelihood estimate. As
opposed to the algorithm described in the last section, where we used
the log-determinant approximation, our theoretical result is
applicable to any strongly convex surrogate for the log-partition
function. The asymptotic analysis presented here is high-dimensional
in nature, i.e. we analyse the estimate in the case when both the
model dimension $p$ and the sample size $n$ tend to
infinity. Traditionally, asymptotic analysis is performed for a fixed
model letting the sample size $n$ to increase, however, that type of
analysis does not reflect the situation that occurs in many real data
sets where the dimensionality $p$ is larger than the sample size $n$
(e.g. gene arrays, fMRIs). To get insight into the behaviour of the
estimate, it is therefore important to perform the high-dimensional
analysis.

The first question to ask is whether the estimate $\hat \theta^n$
converges to $\theta^*$, the true parameter associated with the
distribution? Unfortunately, in general, the answer to this question
is no, since we are using an approximation to the log-partition
function. Note that we could obtain such consistency result if we are
willing to assume that the true graph can be found in a restricted
class of models, e.g. trees for which the approximation will give
the exact solution. 

The next best thing we can hope for is that our procedure produces an
estimate that is close to the best parameter $\hat \theta$ in the
class using the surrogate $B(\theta)$. In general, we cannot tell how
far is the best parameter in the class $\hat \theta$ from the true
$\theta^*$, however, letting the dimension of the model $p$ to
increase with the size of the sample and using a good approximation
$B(\theta)$ we are able to represent an increasing number of
distributions and, hence, reduce the size of the approximation error.

Naturally, the next question is whether the estimate $\hat \theta^n$
at least converges to $\hat \theta$? This convergence would be
obviously true if the model had been fixed, however, we are dealing
with models of increasing dimensionality. In order to guarantee the
closeness of the estimate $\hat \theta^n$ to the best parameter $\hat
\theta$ we will have to assume that the model is sparse and we will
show how the rate of convergence explicitly depends on the
sparsity. We proceed with the main result.

Let $\eta^*$ be the true mean vector corresponding to
$\theta^*$. Since we use strictly convex conjugate dual pair $B$ and
$B^*$, the gradient mapping $\nabla B$ is one-to-one and onto the
relative interior of the constrain set $\OUT(G)$
\cite{rockafellar}. In the limit of infinite amount of data, the
asymptotic value of the parameter estimate is given by $\hat\theta =
\nabla^{-1}B(\eta^*)$. Let $S = \{ \beta\ |\ \hat\theta_\beta \neq
0\}$ be the index set of non zero elements of $\hat\theta$ and let
$\overline{S}$ be its complement. Note that the set $S$ indexes nodes
and edges in the graph and that the size of the set is related to the
sparsity of the estimated graph. Denote $s = |S|$ the number of the
non-zero elements.  The following theorem gives us the asymptotic
behaviour of the parameter estimate $\hat\theta^n$. To prove the
theorem we follow a method of Rothman et al. \cite{rothman08sparse}.
\begin{thm}
Let $\hat\theta^n$ be the minimizer of \eqref{eq:mle_conv}. If
$B(\theta)$ is strongly convex and $\lambda_n \asymp \sqrt{ \frac{
    \log p}{n}}$ then, $$ \norm{\hat\theta^n - \hat\theta}_2 =
\mathcal{O}_P \left(\sqrt{\frac{(p+s)\log p}{n}}\right).$$
\end{thm}
\begin{proof}
 Let $G : \mathbb{R}^d \mapsto \mathbb{R}$ be a map defined as
$G(\delta)  :=  \ell(\hat\theta+\delta; \mathcal{D}) - \lambda^n
\sum_{uv}|\hat\theta_{uv} + \delta_{uv}|  
- \ell(\hat\theta; \mathcal{D}) + \lambda^n
\sum_{uv}|\hat\theta_{uv}|$.
Using the Taylor expansion and the fact that $\nabla B(\hat\theta)
= \eta^*$, we have 
$
B(\hat\theta+\delta) - B(\hat\theta) =
\langle \eta^*, \delta \rangle +
\frac{1}{2}\delta^T\nabla^2[B(\hat\theta+\alpha\delta)]\delta, 
$
for $\alpha \in [0, 1]$. Now, we may write G as:
\begin{equation} \label{eq:rewrite_g}
\begin{aligned}
G(\delta) & = \langle \delta, \hat\eta^n - \eta^* \rangle
    - \frac{1}{2}\delta^T\nabla^2[B(\hat\theta+\alpha\delta)]\delta\\
    & - \lambda^n \sum_{vv'}(|\hat\theta_{vv'} + \delta_{vv'}| -
                        |\hat\theta_{vv'}| ).
\end{aligned}
\end{equation}
By construction, our estimate $\hat\delta^n = \hat\theta^n -
\hat\theta$ maximizes $G$ and we have $G(\hat\delta^n) \geq G(0) =
0$. The proof continues by showing that for some $L > 0$ and
$\norm{\delta}_2 = L$ we have $G(\delta) < 0$, which implies that
$||\hat \delta^n||_2 \leq L$ because of concavity of
$G$. Appropriately choosing $L$, we show that $\delta$ converges to
$0$.

We proceed by bounding each term in \eqref{eq:rewrite_g}. The first
term can be written as:
\begin{equation}
{\small
\begin{aligned}
|\langle \delta, \hat\eta^n - \eta^*\rangle| & \leq
|\sum_{uv \in E} \delta_{uv} (\hat\eta_{uv}^n - \eta_{uv}^*)| +
|\sum_{v \in V} \delta_{v} (\hat\eta_{v}^n - \eta_{v}^*)| \\
& \leq (*) + (**).
\end{aligned}
}
\end{equation}
Using the union sum inequality and Hoeffding's inequality, with
probability tending to $1$, we have
$$ \max_{uv}|\hat\eta_{uv}^n -
\eta_{uv}^*|\ \leq\ C_1 \sqrt{\frac{\log p}{n}},
$$
and a bound $(*) \leq C_1 \sqrt{ \frac{\log p}{n}}
 \sum_{uv}|\delta_{uv}|$. Using the Cauchy-Schwartz inequality and
 Hoeffding's inequality the second term is bounded as 
\begin{equation*}
 {\small
\begin{aligned}
 (**) & \leq (\sum_v(\hat\eta_v^n -
  \eta_v^*)^2)^{1/2}(\sum_v\delta_v^2)^{1/2} \\
&\leq C_2\sqrt{\frac{p \log p}{n}} (\sum_v
   \delta_v^2)^{1/2}.  
\end{aligned}
}
\end{equation*}
The Hessian of a strongly convex function is positive definite, so the
second term in \eqref{eq:rewrite_g} can be bound as follows $ \bm{\tr}
(\delta^T \nabla^2 B(\hat\theta + \alpha\delta) \delta) \geq C_3
\norm{\delta}_2^2$, where $C_3$ is a constant that depends on the
minimum eigenvalue of the Hessian. Using the triangular inequality on
the third term, we have 
$$
\sum_{vv'} (|\hat\theta_{vv'} + \delta_{vv'}|
- |\hat\theta_{vv'}| ) \geq (\sum_{vv' \in \overline{S}}|\delta_{vv'}|
- \sum_{vv' \in S} |\delta_{vv'}|).
$$
Now we can define the constant $L$ as,
$$
L = M\left (\sqrt{ \frac{s\log p}{n} } + \sqrt{ \frac{p\log p}{n}}
\right) \rightarrow 0.
$$ 
Taking $\lambda^n = \frac{C_1}{\epsilon} \sqrt{\frac{\log
    p}{n}}$, we have an upper bound on $G$:
\begin{equation*}
\begin{aligned}
G(\delta) & \leq C_1 \sqrt{ \frac{\log p}{n}}(1-\frac{1}{\epsilon})
\sum_{uv \in \overline{S}}|\delta_{uv}| \\
& + C_1 \sqrt{ \frac{\log
    p}{n}}(1+\frac{1}{\epsilon}) \sum_{uv \in S}|\delta_{uv}| \\
& + C_2\sqrt{\frac{p \log p}{n}} (\sum_v \delta_v^2)^{\frac{1}{2}} - C_3
\norm{\delta}_2^2.
\end{aligned}
\end{equation*}
First term is negative for small $\epsilon$ so we can remove it from
the upper bound. Using the fact that $\sum_{vv' \in S}|\delta_{vv'}|
\leq \sqrt{s}(\sum_{vv' \in S} \delta_{vv'}^2)^{\frac{1}{2}}$, the
upper bound becomes 
$
G(\delta) \leq ( \frac{C_1(1+\epsilon)}{\epsilon
  M} - C_3) (\sum_{vv'}\delta_{vv'}^2 )^{\frac{1}{2}} +
(\frac{C_2}{M} - C_3) (\sum_{v}\delta_{v}^2
)^{\frac{1}{2}} 
< 0
$
for sufficiently large $M$ and the theorem follows.
\end{proof}

\section{Experimental Results}

In this section we compare the performance of our estimation method
for sparse graphs as well as Banerjee et al.~\cite{banerjee07model}
and Wainwright et al.~\cite{Wainwright06High} on simulated data. In
order to assess the performance we compare speed, accuracy of the
structure selection and accuracy of the estimated parameters. Method
of Wainwright et al.~\cite{Wainwright06High} produces two estimates
$\tilde \theta_{uv}$ and $\tilde \theta_{vu}$ for each edge
parameter. There are two ways how we can symmetrize the solution:
$$
\hat \theta_{uv} = \left\{ 
  \begin{array}{ll} 
    \tilde \theta_{uv} & \text{if } |\tilde \theta_{uv}| < |\tilde
    \theta_{vu}|\\
    \tilde \theta_{vu} & \text{if } |\tilde \theta_{uv}| \geq |\tilde
    \theta_{vu}|
  \end{array} 
\right.
\text{  ``Wainwright\_min''},
$$
and
$$
\hat \theta_{uv} = \left\{ 
  \begin{array}{ll} 
    \tilde \theta_{uv} & \text{if } |\tilde \theta_{uv}| > |\tilde
    \theta_{vu}|\\
    \tilde \theta_{vu} & \text{if } |\tilde \theta_{uv}| \leq |\tilde
    \theta_{vu}|
  \end{array} 
\right.
\text{  ``Wainwright\_max''}.
$$
We have implemented the method of Wainwright et
al. \cite{Wainwright06High} using a coordinate descent algorithm for
logistic regression \cite{friedman08regularized}. To compare the
method of Banerjee et al. \cite{banerjee07model} we used ``COVSEL''
package available from the authors website.

We use three types of graphs for experiments: (a) 4-nearest-neighbor
grid models, (b) sparse random graphs, and (c) sparse random graphs
with dense subgraphs. The grid model represents a simple sparse graph,
which does not allow for exact inference due to the large tree
width. To create a random sparse graph, we choose a total number of
edges and add them between random pairs of nodes, taking into an
account the maximum node degree. Random graphs with dense subgraphs
are globally sparse, i.e.~they have few edges, however there are local
subgraphs that are very dense, with strong interactions between
nodes. To generate them, we first create dense components and then
randomly add edges between different components. For a given graph, we
assign each node a parameter $\theta_v \sim \mathcal{U} [-1,1]$ and
each edge a parameter $\theta_{uv} \sim \mathcal{U} [-\xi, \xi]$,
where $\xi$ is the coupling strength. For a given distribution
$\mathbb{P}_{\theta^*}$ of the model we generate random data sets $\{
x^{(1)}, x^{(2)}, \ldots, x^{(n)}\}$ of i.i.d. points using Gibbs
sampling. Every experimental result is averaged over 50 runs.

We first comment on the speed of convergence of the methods. We have
decided not to plot graphs with speed comparisons due to the use of
different programming languages, however, from our limited experience
we observe that the method of Wainwright et
al.~\cite{Wainwright06High} is the fastest and the running time does
not depend much on the underlying sparsity of the graph. Our method
compared favorably to the method of Banerjee et
al.~\cite{banerjee07model}, however, increasing the density of the
graph or increasing the coupling strength $\xi$ resulted in an
increase of the number of added cycle-inequalities needed for the
approximation and in a slower convergence. From the comments on the
speed of the methods, we can suspect that there is a trade-off between
speed and accuracy.

\begin{figure}[tp]
     \centering     
     \includegraphics[width=0.5\textwidth]{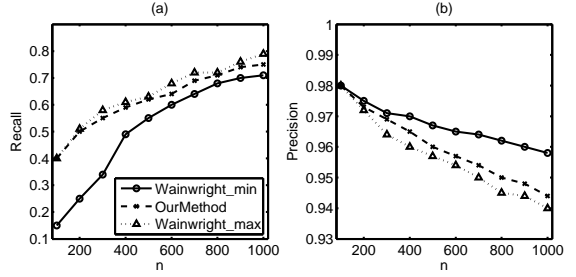}
     \caption{Comparing edge recovery of a sparse random graph:
       $p=50$, $\xi=0.5$, $Nedges=100$.}
     \label{fig:struct_a}
\end{figure}

\begin{figure}[tp]
     \centering     
     \includegraphics[width=0.5\textwidth]{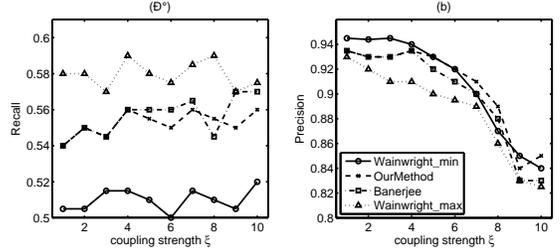} 
     \caption{Comparing edge recovery of a graph with dense subgraph:
       $p=50$, $n=500$, $Nedges=100$. }
     \label{fig:struct_b}
\end{figure}

Next, we compare the accuracy of the edge selection. For this
experiment we create a random sparse graphs with $p=50$ nodes and
$100$ edges, such that the coupling strength is $\xi = 0.5$. Then we
vary the sample size $n$ from $100$ to $1000$. Figure
\ref{fig:struct_a} shows precision and recall of edges included into
the graph for a regularization parameter set as $\lambda_n =
2*\sqrt{\frac{\log p}{n}}$. We have excluded the method of Banerjee et
al. \cite{banerjee07model} in the figure, since for this experiment
the estimated structure was identical. From plots we can see that our
method and ``Wainwright\_min'' produce similar results and perform
better than ``Wainwright\_max''. Similar conclusions can be drawn for
the grid model (results not shown). To shed some more light on these
results, it is instructive to discuss the speed at which the edges get
included into the model as the function of parameter $\lambda_n$.
``Wainwright\_max'' produces estimates that include edges the fastest
and the resulting graph is the densest which can explain low precision
due to many spurious edges that get included into the
model. ``Wainwright\_min'' includes edges most conservatively, while
our solution produces graphs that according to their denseness fall in
between ``Wainwright\_min'' and ``Wainwright\_max'' estimates. Next,
we move onto a harder problem of estimating the structure of graphs
that have strong couplings between nodes. For this experiment, we
construct a random graph with two dense subgraphs, which are fully
connected graphs of size $8$. Graphs have total of $p=50$ nodes and
$100$ edges. We vary the coupling strength $\xi$ in interval $[1,
  10]$. The structure is estimated from a sample size of $n=500$ and
the results are presented in Figure~\ref{fig:struct_b}. In this
experiment we start to notice a difference between the estimated
models using our method and the method \cite{banerjee07model}. As the
coupling strength increases, the mean parameter is not captured within
the constraints defined by the cycle-inequalities and our method has
to add the violated cycle-inequalities to the constraint set $\OUT(G)$
which produces a different estimate.

\begin{figure}[tp]
     \centering     
     \includegraphics[width=0.5\textwidth]{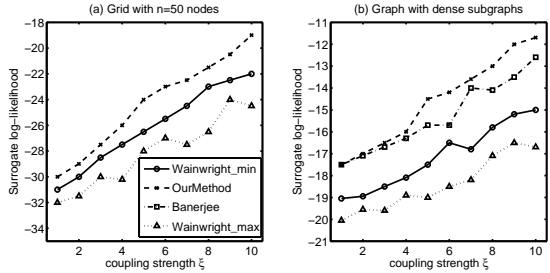} 
     \caption{Test surrogate log-likelihood. $\theta_{uv} \in [-\xi, \xi]$.
       (a) Sparse grid $p=50$, $n=500$;  (b) Graph with dense
       subgraphs $p=50$, $n=200$, $Nedges = 100$. }
     \label{fig:surrogate}
\end{figure}

Final set of experiments measures how well the estimated model fits
the observed data and for that purpose we use surrogate
log-likelihood. While the true measure of the fit should be the
log-likelihood, it is not possible to use it due to the problem of
evaluating the log-partition function. Furthermore, in practise we
always use the surrogate log-likelihood when choosing a model from
data, so our experiments can be justified. Figure \ref{fig:surrogate}
shows the fit for data generated from the grid model. As in estimation
of the structure, for this sparse model, there is no difference
between our method and the method of Banerjee et
al.~\cite{banerjee07model}. One explanation of this result is that the
outer bound on the polytope $\mathcal{M}$, implicitly defined through
log-determinant that act as a log-barrier, captures the estimated mean
parameter and all the cycle-inequalities are satisfied. Next,
similarly to the structure estimation, we generate a graph with dense
subgraphs and present results in Figure \ref{fig:surrogate}. Again, we
can see that our method start to perform better as we increase the
strength of couplings. On Figure \ref{fig:surrogate} we also plot the
fit for parameters estimated from the method of Wainwright et
al.~\cite{Wainwright06High}, however, since the method uses a
different pseudo-likelihood, it performs worse than the other two
methods.

To summarize, we have shown that on the task of estimating the
structure our method performs similarly to the method of Banerjee et
al.~\cite{banerjee07model} and that both methods estimate structure
that falls between ``Wainwright\_min'' and ``Wainwright\_max''
estimates. When measuring how well do learned models fit the data, we
observe that our method outperforms the method of Banerjee et
al.~\cite{banerjee07model} when the model has dense subgraphs and
strong interactions between nodes.

\section{Conclusion}

In the paper we have presented a method for jointly learning the
structure and the parameters of an undirected MRF from the data. Our
method is useful when there are strong correlations between nodes in
the graph or when the true graph is not too sparse. If the true graph
is very sparse, our algorithm efficiently finds the estimate without
performing the cutting-plane step. We showed how to incorporate the
class of cycle-inequalities into the algorithm, which are particularly
valuable as they can be separated efficiently and added as needed to
improve the solution. Furthermore, our algorithm can be efficiently
combined with the algorithm for projection onto the $\ell_1$ ball in
cases when the parameters are constrained to lie in the $\ell_1$ ball.

We have analyzed convergence rate of an estimate based on maximizing
a penalized surrogate likelihood in high-dimensional settings. We have
given a rate that depends on the number of non-zero elements of the
best parameter for the surrogate likelihood. Such analysis provides
insight into the performance of the method in high-dimensional
setting, when both $p$ and $n$ are allowed to grow.

\section{Acknowledgements}

This material is based upon work supported by an NSF CAREER Award to EPX
under grant No. DBI-0546594, and NSF grant IIS-0713379. EPX is also
supported by an Alfred P. Sloan Research Fellowship of Computer
Science.

{\small
\bibliography{cutting_plane}
}

\end{document}